# Disentangling Codemixing in Chats: The NUS ABC Codemixed Corpus

**Svetlana Churina, Akshat Gupta, Insyirah Mujtahid** and **Kokil Jaidka**
Centre for Trusted Internet & Community,
National University of Singapore,
Singapore

## Abstract

Code-mixing involves the seamless integration of linguistic elements from multiple languages within a single discourse, reflecting natural multilingual communication patterns. Despite its prominence in informal interactions such as social media, chat messages and instant-messaging exchanges, there has been a lack of publicly available corpora that are author-labeled and suitable for modeling human conversations and relationships. This study introduces the first labeled and general-purpose corpus for understanding code-mixing in context while maintaining rigorous privacy and ethical standards. Our live project will continuously gather, verify, and integrate code-mixed messages into a structured dataset released in JSON format, accompanied by detailed metadata and linguistic statistics. To date, it includes over 355,641 messages spanning various code-mixing patterns, with a primary focus on English, Mandarin, and other languages. We expect the Codemix Corpus to serve as a foundational dataset for research in computational linguistics, sociolinguistics, and NLP applications. Code and dataset sample can be found here.

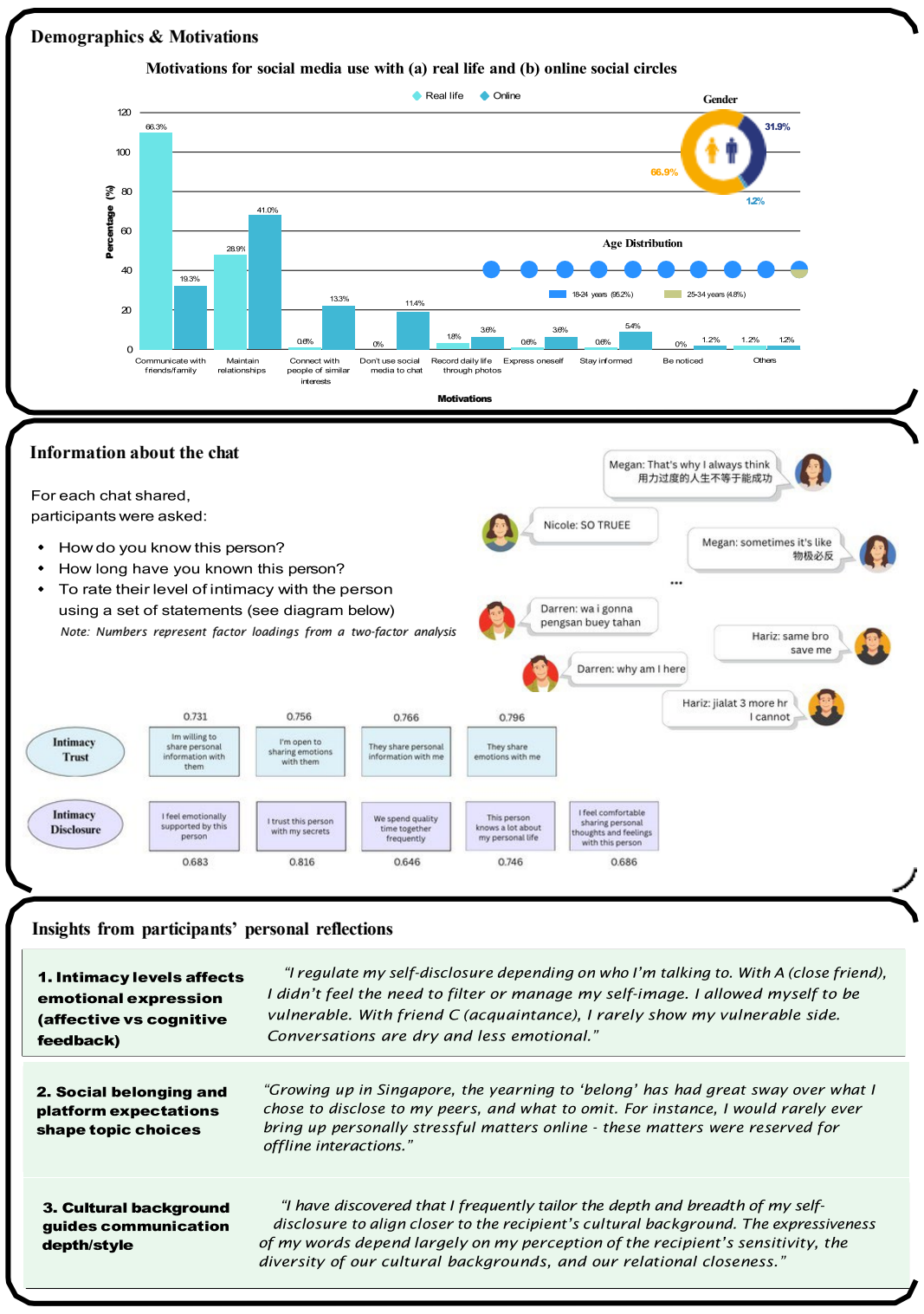

Figure 1: Top: Information collected about survey participants. Middle: Sample chat snippets illustrating varying tones and relational dynamics. Participants rated their chats on intimacy, using items grouped under two dimensions—*Trust* and *Disclosure*. Bottom: Insights from participants' personal reflections on communication practices.

## 1 Introduction

Instant messaging has become a primary mode of communication in the digital era, widely adopted across contexts—from personal interactions with family and friends to professional and educational exchanges. Unlike social media platforms, which often encourage performative and public-facing content, instant messaging offers a private, real-time, and often more intimate channel for interaction. While numerous studies have examined the role of emojis in digital communication (Boutet et al., 2021; Erle et al., 2022; Riordan, 2017), there is still a lack of research exploring how communication—particularly code-mixing—varies with the depth of interpersonal relationships and levels of intimacy. Understanding this relationship is crucial for capturing the nuanced ways people express closeness, familiarity, or social distance in private conversations.

In this study, we introduce the Air Batu Campur (ABC) Codemixed Corpus - a textured dessert consisting of Southeast Asian ingredients, and also a live, publicly available corpus of instant messages, written mostly in English and Southeast Asian languages, collected through voluntary data donation. Participants were asked to share private conversa-tions with three different partners, each representing varying levels of relational intimacy. Our data donation framework is inspired from previous efforts to collate an instant messaging dataset (Chen and Kan, 2013), which we extended by collecting donor responses to standard psychometric scales. It balances research value with participant privacy: it includes instructions to consenting donors to self-anonymize their sensitive content and affords them control over the data shared. This methodology sets a replicable standard for ethically collecting real-world, private conversational data at scale.

This paper provides a descriptive analysis of the dataset, reporting the distribution of languages, message lengths, and code-mixing rates across different levels of relational intimacy. A key focus of the dataset is the inclusion of code-mixed messages, enabling researchers to explore how language use—particularly code-mixing—varies across intimacy levels. It is also a valuable resource for studying the nuanced dynamics of private communication and the sociolinguistic factors that influence code-mixing behavior.

## 2 Background

Southeast Asia is one of the most ethnically and linguistically diverse regions in the world (Leng, 1980). Each country in the region is internally diverse, varying in ethnicity, language, and religion (Takagi, 2009). Over centuries, cultural fluidity and migration patterns have shaped the region's linguistic landscape, giving rise to multiracial and multicultural societies where multiple languages coexist. However, while linguistic diversity remains a defining feature of the region, globalization has also contributed to the gradual assimilation of vernacular languages, as dominant languages, such as English, take center stage in everyday communication (Singh et al., 2012; Ramli et al., 2021).

English has emerged as a key language in Southeast Asia due to its association with upward mobility, economic opportunity, and global interconnectedness (Albury, 2016). Its dominance has been driven largely by education policies across the region that position English as a critical tool for national development (Sercombe, 2019).

In Singapore, this linguistic shift was formalized in 1996, when the government introduced bilingual education policies, making English the primary language of instruction while requiring students to learn their respective mother tongues (Leimgruber, 2013). These policies not only elevated English as a dominant language in formal communication but also contributed to a unique linguistic environment where diverse languages and dialects continued to thrive in informal settings (Gupta, 1989) . Given the country's rich linguistic diversity, the practice of code-mixing emerged as a natural consequence, with speakers fluidly blending multiple languages in everyday conversations.

As a result of this widespread code-mixing, Singlish, Singapore's colloquial variety of English, emerged as a distinct vernacular, blending elements from Mandarin, Hokkien, Cantonese, Malay, and other languages. Widely used in informal settings, Singlish reflects the nation's multicultural identity and serves as a linguistic marker of local belonging. This phenomenon arises from the tension between 'being global' and 'being local' (Alsagoff, 2007), shaping distinctive communication styles that balance global influences with cultural heritage. In Singapore, code-mixing serves several key communicative functions, including signaling group identity (Kipchoge, 2024), facilitating communication (Sumartono and Tan, 2018), and establishing rapport (Bolton and Botha, 2019). Across Southeast Asia, similar patterns of code-mixing have emerged, reflecting its role as a strategic tool for various communicative functions.

## 3 Prior Work

While code-mixing has gained attention, few large-scale, high-quality datasets capture its real-world complexity. Existing resources often focus on a single language pair, lack speaker metadata, or are limited to scripted or social media content. For example, L3Cube-HingCorpus (Nayak and Joshi, 2022) offers 52 million Hindi-English tweets but lacks conversational context. Bollywood Romanized Corpus (Khanuja et al., 2020) draws from movie scripts, not spontaneous speech. In Southeast Asia, Singapore's NSC (Koh et al., 2019) provides read speech data with local lexical items but not code-mixing in personal communication. CoSEM (Gonzales et al., 2023) and the NUS-SMS Corpus (Chen and Kan, 2013) include demographic metadata, while (Foo and Ng, 2024) disambiguate codemixed 'Singlish' (Singaporean English) discourse particles for machine translation. However, these studies and datasets are small in size and lack any annotation of relationship dynamics between participants. BOLT Phase 2 (Song et al., 2014) cap-tures informal multilingual chat data, but primarily focuses on monolingual exchanges. TweetTaglish (Herrera et al., 2022) adds Tagalog-English tweets but with some collection noise.

Table 1: Corpus Composition and Message Statistics

| Part A: Overall Corpus Statistics | |
|---|---|
| **Metric** | **Value** |
| Total Messages Collected | 355,641 |
| Unique Contributors | 154 |
| Languages Represented | English, Mandarin, Hokkien, Tamil, Malay, Japanese, Korean |
| Average Message Length | 4.3 tokens |
| Code-Mixed Messages (%) | 22% |

| Part B: Message Statistics by Intimacy Level | | | |
|---|---|---|---|
| **Statistic** | **Acquaintances** | **Frequent Chatting** | **Daily Interactions** |
| Total Messages | 14,166 | 44,590 | 280,745 |
| Mean Length of Messages (Words) | 6.82 | 4.59 | 4.12 |
| Median Length of Messages (Words) | 5.0 | 3.0 | 3.0 |
| Standard Deviation of Message Length | 9.51 | 33.42 | 5.36 |
| Code-Mixing (%) | 7.42% | 12.26% | 19.25% |
| Transliteration (%) | 2.21% | 4.16% | 0.75% |

Our dataset fills this gap by providing authentic, multilingual private messages from Singaporeans across different relationship types. It includes demographics, relational context, and naturally occurring code-mixing involving English, Chinese, Malay, and Tamil—going beyond the typical two-language focus. This opens new challenges and opportunities for studying real-world multilingual communication.

## 4 Methodology

We collected 477 chat threads over six months from Singaporean university students who responded to a university-wide recruitment call. Participants were provided with definitions and examples to distinguish code-mixing from code-switching and asked to submit relevant conversations, covering different levels of relationship intimacy. For each submission, participants provided metadata such as the closeness of the relationship, length of acquaintance, and perceived intimacy. To contextualize code-mixing practices, participants also completed surveys on their demographics and social media use. To incentivize follow-up, participants could optionally provide an email address to receive a $5 food delivery voucher; these addresses were securely discarded after distribution. Participants under 18 were excluded. Out of 332 recruits, 166 did not contribute any data.

## 5 Corpus Statistics and Analysis

Of the 166 contributors, 111 identified as female, 53 as male, and 2 preferred not to disclose their gender (Figure 1). All participants were at least 18 years old, with most (158) aged 18–24, and a smaller group (8) aged 25–34. Telegram and WhatsApp were the most used platforms, with 165 and 163 users, respectively. Fewer participants reported using WeChat (18) or Discord (8), while Facebook and Snapchat were rarely used.

The dataset comprises 355,641 messages, with an average length of 4.3 words (median: 3), reflecting the brevity typical of digital messaging. Messages were categorized based on relationship type into three groups:

- **Close Relationships**: family members, close friends, partners
- **General Relationships**: friends, acquaintances, community members
- **Work Relationships**: classmates, colleagues, coworkers

As shown in Table 1, messages tended to be shorter in closer relationships. Messaging patterns also reflected social motivations: most participants used messaging apps to stay connected with friends and family (110) or maintain relationships (48), both offline and online. A smaller group (20) used them to meet new people with shared interests, while others used them for self-expression (4) or daily life documentation (6). These trends are summarized in Figure 1.

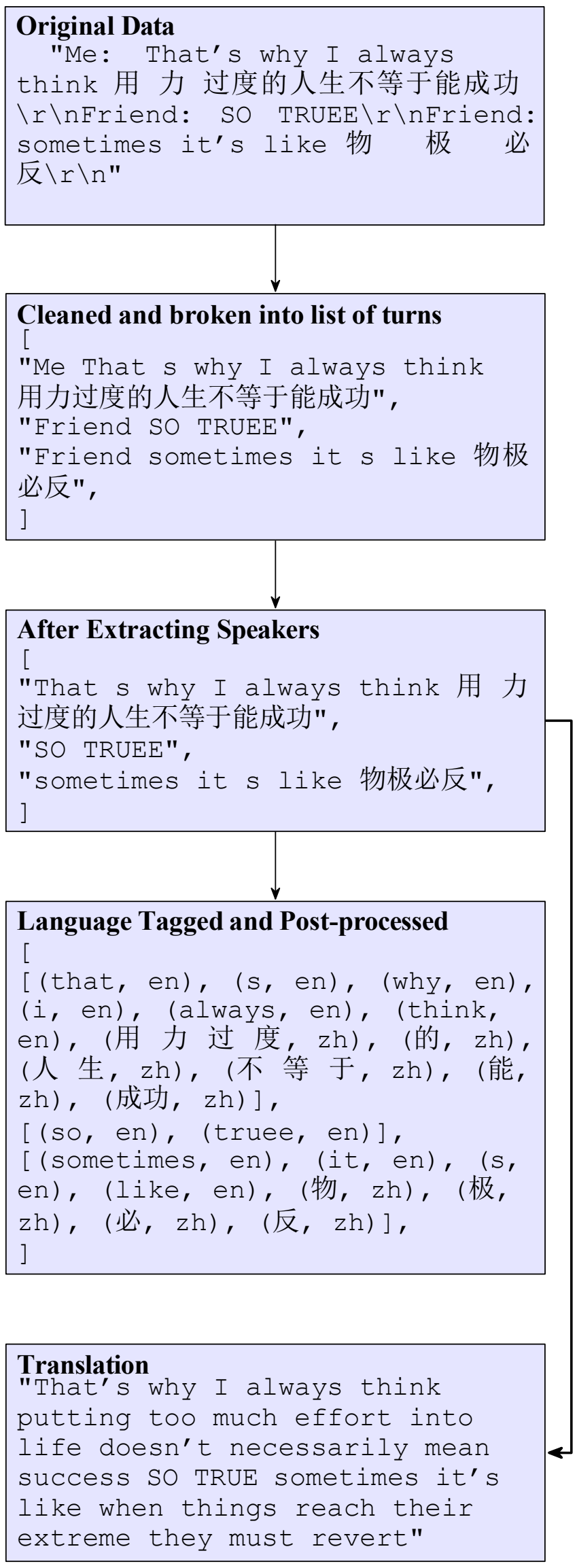

Figure 2: Language tagging and translation pipeline (en = English, zh = Mandarin).

## 6 Data Preprocessing, Language Identification, and Translation

The dataset consists of user-generated messages collected from WhatsApp and Telegram, each containing metadata such as the sender ID, timestamp, and conversation ID. The messages are presented in chronological order within conversation threads. Each message is treated as a single text unit, which may contain code-mixed content spanning two or more languages.

We applied several preprocessing steps to clean the chats, as illustrated in Figure 2, including the removal of URLs, emails, file extensions, timestamps, and excess punctuation. Emojis, numbers, and casing were preserved to retain linguistic signals useful for identifying proper nouns and code-mixing behavior. Conversations were segmented into message turns using newline or > characters as boundaries, and speaker names were extracted using frequent word patterns where available. Some participants donated the same chats across multiple intimacy levels. To maintain data integrity, we excluded these repeated chats from the dataset due to the potential for data redundancy. The comprehensive pre-processing details are provided in the Appendix.

For language identification, we combined rule-based methods (NLTK's `WordNet` and `wordfreq`) with model-based approaches, including fastText's pretrained model (Joulin et al., 2017) and XLM-RoBERTa (Papluca, 2020). Given their limitations in handling informal Southeast Asian code-mixed texts, we further applied token-level tagging using Qwen2.5-14B-Instruct (Hui et al., 2024), prompting it specifically for Southeast Asian languages and online texting patterns. Post-processing aligned tags with original words, normalized language labels, and identified scripts (e.g., Roman, Chinese). Full pipeline specifications and prompts are detailed in the Appendix.

The entire code-mixed corpus was also translated into English for downstream analysis using the Qwen2.5-14B-Instruct model, due to its contextual understanding and multilingual capabilities (Hui et al., 2024). Initially, translations were attempted at the message level, but the model often indicated insufficient context. Therefore, we translated at the conversation level instead, with longer threads chunked as needed to maintain context. The complete translation prompt is included in the Appendix. Post-processing steps corrected common issues such as repeated words, joined tokens, and untranslated segments.

### 6.1 Tagging Performance

For evaluation purposes, we created a stratified random sample of 100 conversations, based on the presence of the most frequent languages, which together cover over 90% of all tokens. On this sample, we compare the language tags identified by our primary approach (Qwen2.5) to two reference methods (fastText and papluca/xlm-r). We computed Cohen's Kappa and Tag agreement rate between tags from our primary approach and the reference methods. As shown in Table 2, while there is a fair agreement in the output labels overall, the low Cohen's Kappa indicates that the agreement between labels might be only limited to the majority labels like English. Also, since LLMs are prone to paraphrasing, the number of output tokens are not the same as the baseline methods which have a determined number of tokens. Thus, the comparison was done on cases where the number of output tokens was the same (n = 43).

| Baseline | Cohen's Kappa | Agreement Rate |
|---|---|---|
| **fastText** | 0.25 | 0.71 |
| **papluca/XLM-RoBERTa** | 0.26 | 0.65 |

Table 2: Tag agreement – Qwen2.5 primary tags vs. baseline methods (on 43 conversations).

| Original | Baseline_FT | Baseline_XLMR | Qwen2.5 |
|---|---|---|---|
| lemme try | `(lemme,sl), (try,en)` | `(lemme,it), (try,en)` | `(lemme,en), (try,en)` |
| nihao i just had dinner | `(nihao,en), (i,en), (just,en), (had,en), (dinner,eng)` | `(nihao,sw), (i,en), (just,en), (had,en), (dinner,eng)` | `(nihao,zh), (i,en), (just,en), (had,en), (dinner,eng)` |
| 我 想 要 脱 离 nus | `(我,zh), (想 要,zh), (脱 离,jp), (nus,vi)` | `(我,zh), (想 要,zh), (脱 离,zh), (nus,hi)` | `(我,zh), (想,zh), (要,zh), (脱 离,zh), (nus,propernoun)` |

Table 3: Token-level language tags from Qwen2.5 compared to baseline methods.Codes: *zh* = Mandarin, *jp* = Japanese, *en* = English, *sw* = Swahili, *sl* = Slovenian, *vi* = Vietnamese, *hi* = Hindi, *it* = Italian.

As shown in Table 6.1, LLM-based tagging showed superior performance compared to traditional methods:

- It outperformed other approaches at identifying irregular English words like abbreviation and slang, essential for our dataset.
- It was better at identifying transliterated words, which are heavily used while texting.
- It was able to identify proper nouns from the added context, something which the token-level baseline models are not capable of.

In spite of these advantages, identifying languages within a code-mixed text is not trivial and our approach was not foolproof. As can be seen from Figure 3, tagging within Asian languages still has some issues, which are discussed in the Limitations section.

### 6.2 Translation Performance

For evaluating translations, we used the same stratified sample employed for assessing language tags. We compared our translations to two baselines: one generated using the Google Translate API and the other using Facebook's NLLB-200 MT model.

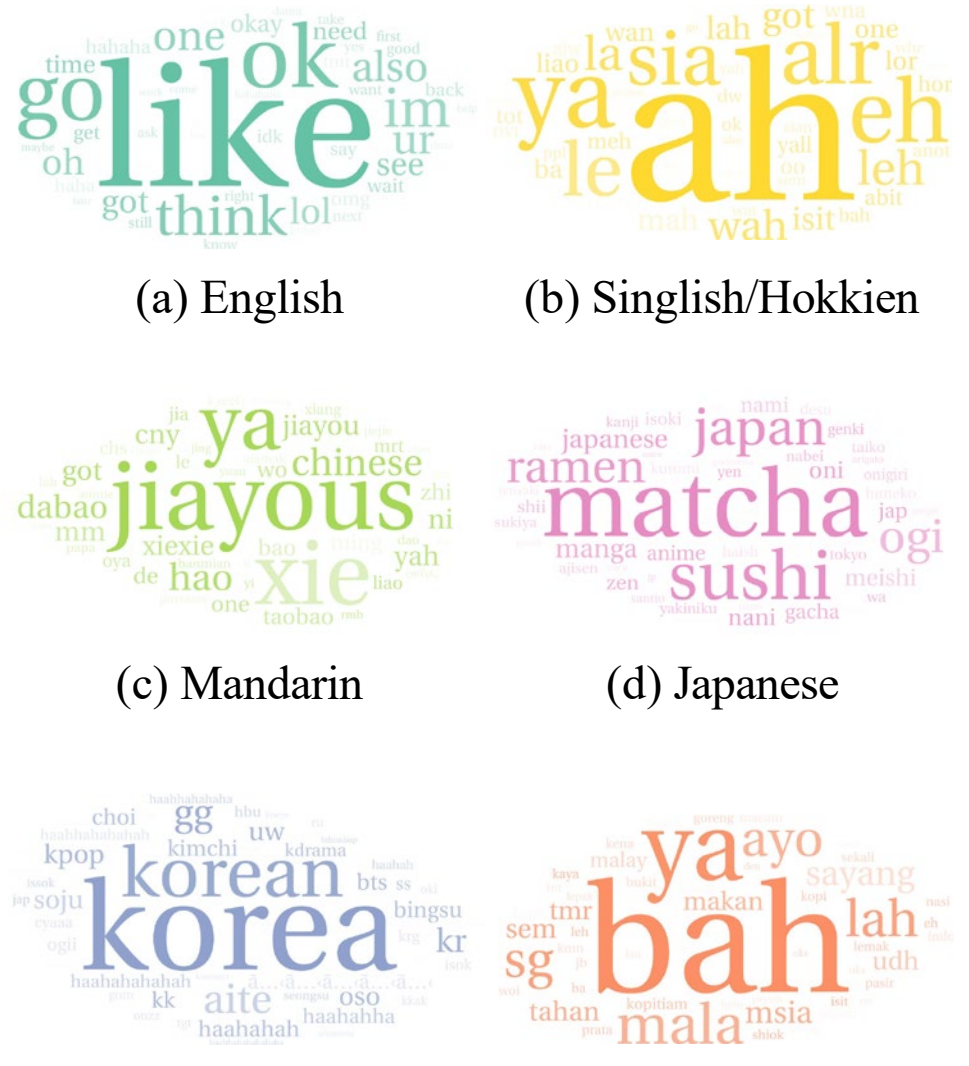

(a) English (b) Singlish/Hokkien

(c) Mandarin (d) Japanese

(e) Korean (f) Malay

Figure 3: Top code-mixed words across six language communities.

| Baseline | BLEU | BERT_F1 | ROUGE-1_F | ROUGE-L_F |
|---|---|---|---|---|
| **Google Translate** | 4.55 | 0.87 | 0.49 | 0.46 |
| **NLLB-200** | 4.47 | 0.81 | 0.40 | 0.35 |

Table 4: Similarity scores of baseline translations compared to Qwen2.5 outputs.

Both baseline translations were performed at the message level, and a single aggregated output was created for comparison. This was necessary due to the limited input handling capabilities of the two baseline models. The similarity scores for our LLM-based translations with the two baselines are shown in Table 4. We made the following observations about the translation performance:

- High BERT scores indicate strong semantic similarity with the baselines, while low BLEU scores reflect lexical divergence—likely due to paraphrasing, as seen in Table 5.
- The LLM-based approach was much better at translating transliterated words, also highlighted in red in Table 5.

## 7 Content Analysis

We conducted a three-level analysis to examine how code-mixing functions across different relationship contexts in Singaporean digital conversations. The detailed analytical framework and our observations are reported in provided in the Appendix, while the following paragraphs offer a broad overview of the findings.

At **Level 1**, we quantified the frequency of code-mixing by relationship type and message length. Nearly all participants used English, often combined with Chinese, Malay, or Tamil. Over 30% of messages contained code-mixing, with usage varying significantly across both relationship types and message lengths (Table 1). Our analysis reveals that code-mixing is most prevalent in shorter messages and in close relationships. As shown in Figure 4, short messages (0–5 tokens) exhibit the highest rates of code-mixing, reaching 36.7% in intimate conversations. The rate decreases with increasing message length, reaching its lowest in the 51–100 token range, before rising again for very long messages (100+ tokens). Interestingly, even in professional or work-related exchanges, short messages show non-trivial levels of code-mixing (about 13%), suggesting that brief conversational turns are more prone to stylistic mixing regardless of social distance.

| Original | Baseline_NLLB | Baseline_GT | Qwen2.5 |
|---|---|---|---|
| Omg jiayouz he said that abt my prev article also I got his approval after the second article | Omg Jiayouz he said that abt my preview article also i got his approval after the second article | Omg jiayouz he said that abt my prev article also I got his approval after the second article | Omg cheer up he mentioned that about my previous article too I received his approval after the second article |

Table 5: Comparison of Qwen2.5 translations with baselines. Red and blue highlight code-mixed and informal words.

At **Level 2**, we examined linguistic features of code-mixed messages using principal component analysis and LIWC-22 (Boyd and Pennebaker, 2022), a text-analysis software that quantifies psycholinguistic features from a text corpus. Code-mixing was positively associated with conversational and relational language, including function words ($r = .28$), pronouns ($r = .17$), and discourse markers. These features suggest code-mixing aligns with social and affective communication goals (Figure 5). We found that structural markers (pronouns, function words), expressive devices (punctuation), and social references are reliable linguistic correlates of code-mixed discourse. We also observed that:

- **Linguistic Structure**: Features such as linguistic ($r = 0.26$, $p_{adj} < .001$), function words ($r = 0.28$, $p_{adj} < .001$), and auxiliary verbs ($r = 0.19$, $p_{adj} < .001$) indicate that code-mixed messages tend to use highly interactive and relational language, characterized by pronouns, auxiliary constructions, and discourse particles typical of casual, present-focused conversation.
- **Social Referencing**: The positive associations with pronoun usage (*pronoun*, $r = 0.17$, $p_{adj} = .009$) and female references (*female*, $r = 0.17$, $p_{adj} = .005$) suggest that code-mixing is more frequent in socially engaging discourse, particularly when referencing other people or relational identities.
- **Contextual Framing**: The use of articles (*article*, $r = 0.16$, $p_{adj} < .01$) and determiners (*det*, $r = 0.23$, $p_{adj} < .001$) highlights the role of referential framing, where speakers specify or differentiate entities in context, potentially to manage clarity in multilingual interaction.

Together, these findings suggest that code-mixing is both, a structural feature of multilingual messaging, and a stylistic resource that aligns with conversational goals such as managing social relationships, expressing affect, and negotiating meaning. In the next section, we extend this analysis by applying the Systemic Functional Linguistics (SFL) framework to interpret how these linguistic patterns contribute to broader conversational strategies and social action.

At **Level 3**, we applied the SFL framework (Halliday and Matthiessen, 2013) to map how code-mixing supports conversational strategies, summarized in Table 6. These include framing messages (e.g., *Actually ah, anyone looking for job?*), expressing emotion (e.g., *Wah I really cannot tahan*), managing closeness (e.g., *Sayang u so much*, *Paiseh we late*), softening tone (e.g., *U da best leh*), and adding humor or emphasis (e.g., *Walao I walked the other way sia*). These examples illustrate how code-mixing helps manage meaning, tone, and relationships in chat interactions.

Building on Elena (2016)'s conceptualization of personal and functional tenor, we also found that speakers strategically modulate tone and style depending on the nature of the relationship and their communicative goals. We outline three tenor-based strategies—affective, interpersonal-social, and discourse-stylistic—that speakers use to manage tone, emotion, and social alignment in digital conversations. These strategies reflect how language choices are shaped by relationship closeness, cultural background, and communicative goals (see Figure 1).

- **Affective Tenor:** Speakers use emotionally

| SFL Dimension | Strategy | Indicators | Caption Example | Type of Relationship |
|---|---|---|---|---|
| **Field**: What the discourse is about.<br>• Topic, content, and activity.<br>• Expressing stress, giving explanations<br>• Referencing shared experiences. | Framing, elaboration, and justification to guide interpretation, explain reasoning, or clarify intent (e.g., *hor, ah, leh*) | hor, ah, leh, meh, bah | "Earlier clean better bah", "Actually ah, since we're talking - you know anyone looking for job?" | Casual friendship or professional acquaintance |
| | Indexing emotional or situational context, including emotional intensity and mood (e.g., *wah, jialat, pengsan, sian*) | wah, jialat, pengsan, relax lah, sian, nua, rabaak, kenna | "Wah I really cannot tahan school already", "I'm too nua sia" | Casual and close friendships |
| **Tenor**: Who is involved.<br>• Participants and relationships.<br>• Roles, social distance, power dynamics. | Managing closeness, face, and solidarity, including affection and politeness (e.g., *sayang, jiayou, paiseh*) | sayang, thanku syg, should be ok one, don't worry la, jiayou, don't pengsan, gamxia, paiseh | "Sayang u so much", "Jiayou, you can do this!", "Should be okay one la", "Hehe paiseh we a bit late", "Gam xia for letting me know" | Close friendship |
| | Constructing identity and relational softening (e.g., *leh, laa, bah, eh*) | leh, laa, liao, bah, eh | "U da best leh", "Thanks so much eh", "U update me bah", "can liao" | Casual friendship |
| **Mode**: How the conversation unfolds.<br>• Medium, channel, and organization.<br>• Written, spoken, or multimodal means. | Managing turns, coherence, and topic shifts (e.g., *ah, hor, eh, liao*) | ah, hor, eh, liao | "Wait ah I think I know why", "Actually hor, I also wanted to check with you" | Casual friendship |
| | Typographic play and exaggerated expressions for humor or alignment (e.g., *walao, nani, gwenchana, siao ah?*) | walao, nani, gwenchana, siao ah? | "HAHAH WALAO I walked the other way sia", "wait nani", "its gwenchana!" | Casual and close friendships |

Table 6: Message-based conversation strategies categorized by SFL dimensions: Field, Tenor, and Mode.

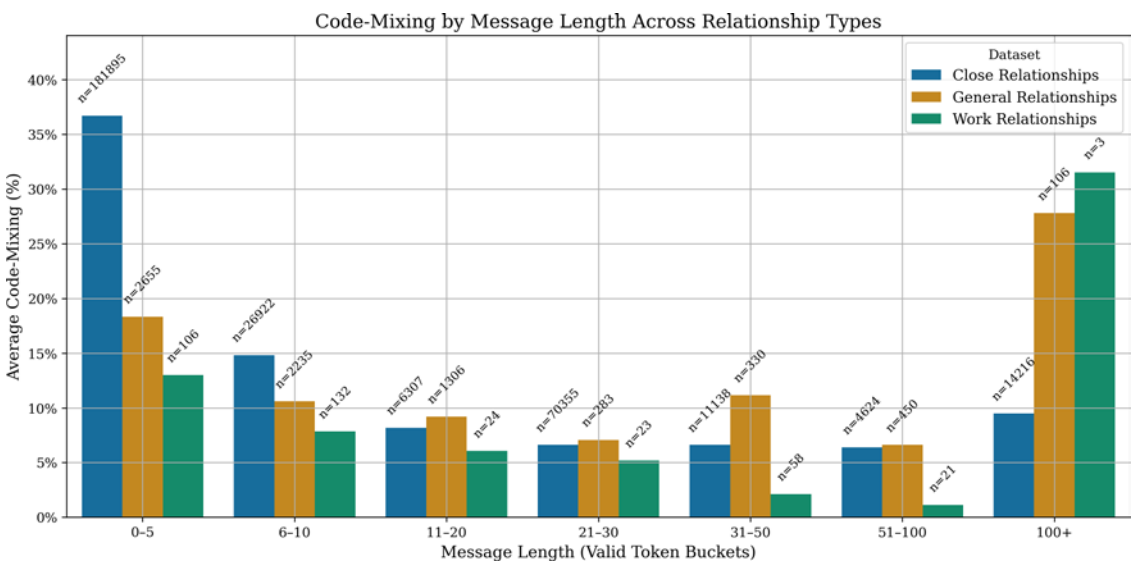

Figure 4: Average percentage of code-mixing across message length buckets (based on valid token count), grouped by relationship type. *n* indicates the number of messages in each bucket per dataset.

charged expressions (e.g., *sayang, jiayou, wah, sian*) to convey care, dramatize feelings, or lighten the mood, especially in close relationships where emotional disclosure is common.

- **Interpersonal-Social Tenor:** Speakers maintain rapport and manage social distance through politeness markers (e.g., *don't worry la, paiseh*) and discourse particles (e.g., *leh, liao*), softening interaction while signaling cultural familiarity and group belonging.
- **Discourse-Stylistic Tenor:** Speakers coordinate conversation flow and frame utterances using particles (e.g., *ah, hor, meh*) and stylized expressions (e.g., *walao, siao ah*) to manage topic shifts, dramatize reactions, and express in-group identity through playful language.

## 8 Discussion and Conclusion

While social media data is increasingly accessible through public APIs or web scraping mechanisms, instant message data remains largely private and inaccessible. Due to the inherently personal nature of instant messaging, such data can only be obtained through voluntary donation by users, posing significant challenges for large-scale research and raising important ethical considerations.

Given these considerations, we believe that the ABC Codemixed Corpus can strengthen multilingual NLP systems, improving both retrieval and classification tasks. For example, Lee et al. (2024) demonstrate that dictionary-guided code-mixed fine-tuning turns an English-centric LLAMA-7B into a stronger multilingual QA model, boosting accuracy by 6–8 percentage points (pp) on low-resource benchmarks like MLQA and XQuAD.

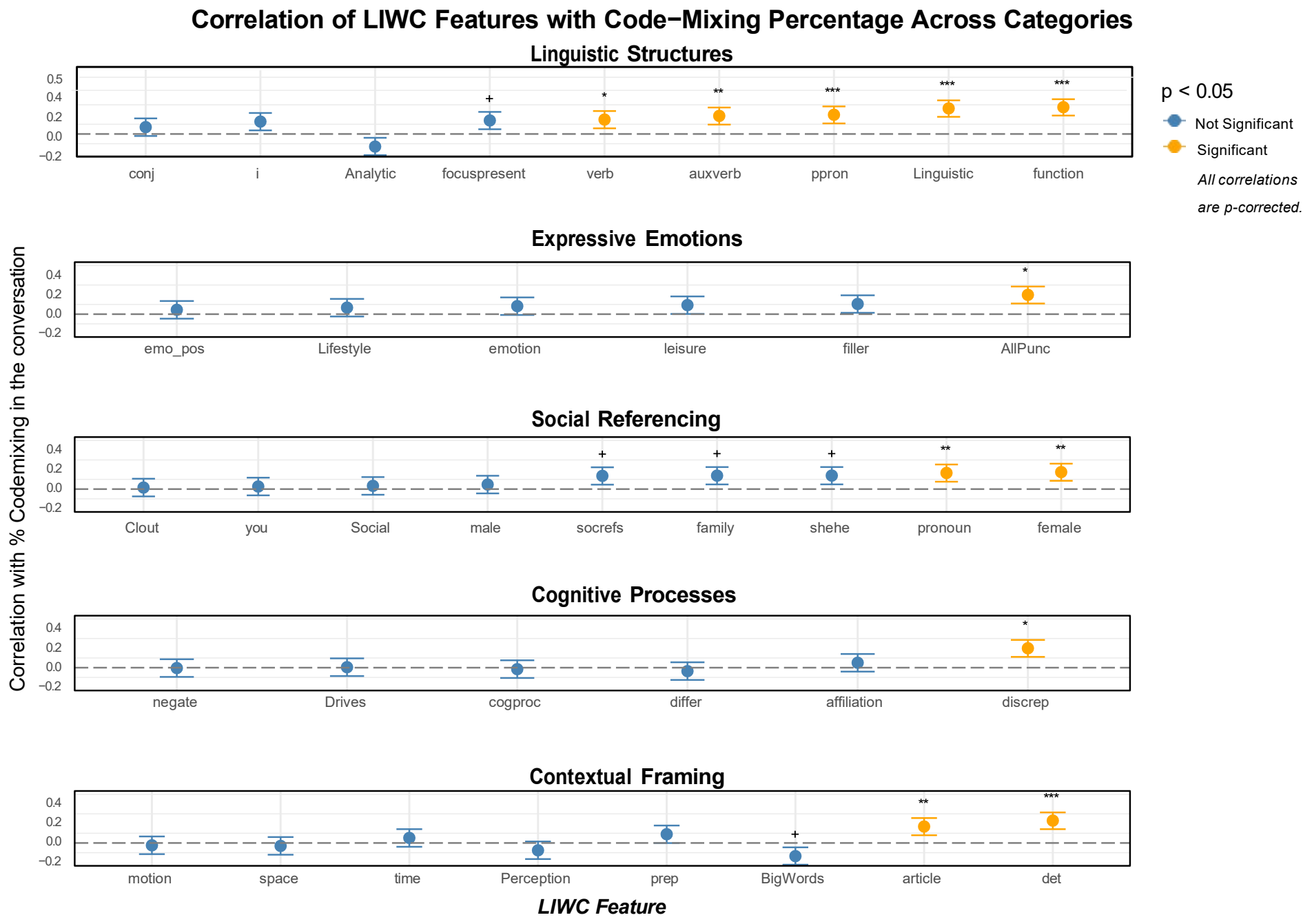

Figure 5: Correlation of LIWC features with the percentage of code-mixed tokens, grouped by five latent linguistic categories derived from principal component analysis. Orange points mark statistically significant p-corrected correlations ($p < .05$).

Similarly, A. Fondekar et al. (2024) introduce FauxHate, a 14,000-tweet Hinglish corpus for hate speech detection. A HingRoBERTa model fine-tuned on this data achieves macro-F1 scores of 0.82 (fake/real) and 0.77 (hate/non-hate), outperforming multilingual RoBERTa and zero-shot LLM prompting by over 6 pp.

Prompt-based methods also benefit from light in-context code-mixing. Shankar et al. (2024) show that inserting code-mixed tokens into few-shot exemplars improves GPT-4's performance by 1.5–6.7 pp on tasks like Telugu disfluency correction and Turkish grammatical-error correction.

However, generation quality and evaluation remain open challenges. Yong et al. (2023) find that while ChatGPT produces more natural mixtures than other LLMs across seven Southeast Asian language pairs, the output remains inconsistent and heavily prompt-dependent. Human ratings of naturalness show wide variability (Fleiss' = 0.216–0.827), reflecting unresolved evaluation issues. Similarly, Zhang et al. (2023) report that prompt-only approaches underperform on several code-switched tasks, including translation, sentiment analysis, and language ID—achieving BLEU scores up to 9 pp lower than supervised baselines. Taken together, these findings suggest that while code-mixed training and prompting can improve model performance, advances in generation control, evaluation methods, and curated resources are needed to realize their full potential.

## 9 Limitations

This study offers an in-depth look at code-mixing in digitally mediated conversations, but several methodological and analytical constraints should be noted.

A primary limitation lies in the scope and representativeness of the dataset. All data were sourced from university students in Singapore, a demographically narrow group whose linguistic practices may differ from those of older adults, working professionals, or speakers in other multilingual communities. Consequently, the findings may not generalize beyond the specific sociolinguistic and cultural context of our participants.

Second, participants submitted selected excerpts rather than full-length conversations. While this approach was necessary to protect privacy and reduce data volume, it limits our ability to analyze turn-taking, conversational flow, or shifts in tone over extended interactions. Some pragmatic or relational cues may therefore be misinterpreted or missed entirely.

Third, although our linguistic analyses captured meaningful associations between code-mixing and communicative strategies, we cannot assume intentionality behind every instance of mixing. Habitual or stylistic language use may not always reflect deliberate social signaling. Future studies incorporating interviews or follow-up elicitation could help clarify user intent.

Fourth, while we used state-of-the-art tools for language identification and tagging, our analysis remains limited by the capabilities of existing models. Complex or low-resource language varieties, such as Singlish or dialectal insertions, may still be misclassified or underrepresented. We plan to explore more targeted approaches in future work, including lexicon-guided correction and dialect-specific adaptation.

## 10 Ethical Considerations

Designing a study involving human subjects, particularly undergraduate student participants, required us to be especially considerate of the ethical implications of our work. To protect personally identifiable information (PII), we implemented a multi-step anonymization process. Messages were pre-screened to remove metadata such as names, email addresses, phone numbers, and file attachments. Nonetheless, due to the informal and varied nature of chat language, we acknowledge the possibility that some PII may persist. We advise dataset users to conduct additional anonymization as needed and to report any identified issues to the authors. Optional email addresses were collected solely for participant compensation (via food delivery vouchers) and securely discarded after disbursement. No personally identifying metadata is included in the released dataset.

We also recognize that participants' self-reported information (e.g., relationship closeness or emotional tone) may be subject to bias or inaccuracy. Furthermore, because participants contributed selected chat excerpts rather than full conversations, some messages may lack context, potentially affecting the interpretation of tone, intent, or social dynamics.

Finally, our dataset reflects the linguistic behavior of a relatively homogeneous group—Singaporean university students—which may limit generalizability. However, the inclusion of participants' personal reflections, alongside linguistic data, provides valuable insight into motivations and communication strategies in multilingual digital settings. This corpus is intended strictly for academic use under a research-only license.

**Acknowledgment:** AI was used to proofread a draft of this paper and format the tables.

# A Datasheet for the ABC Codemix Dataset

## A.1 Motivation

**For what purpose was the dataset created?**
The ABC Codemix Dataset have been created with purpose to study how private message communication varies across different levels of interpersonal intimacy, with a particular focus on the use of code-mixing.

**Who created the dataset?**
The dataset was curated and annotated by Anonymous.

**Who funded the creation of the dataset?**
This work was supported by Anonymous.

## A.2 Composition

**What do the instances that comprise the dataset represent?**

The dataset consists of individual messages taken from private instant messaging conversations. Each row is enriched with information about the participants, including their demographics and the type of relationship they share.

**How many instances are there in total?**
The dataset consists of 477 donated chats.

**Does the dataset contain all possible instances?**
No, dataset does not contain all possible messages exchanged between participants. Instead, participants voluntarily donated selected portions of their private messages that they felt comfortable to share.

**What data does each instance consist of?**
Each instance consists of a message (or set of messages) from a private instant messaging conversation, along with metadata about the participants and their relationship. This includes the message text, code-mixing information, participant demographics (age and gender), self-reported intimacy scores, relationship category, and details about their messaging app usage and communication motivations.

**Is there a label or target associated with each instance?**
Yes, each instance is associated with a relationship category level, as well as self-reported intimacy scores.

**Is any information missing from individual instances?**
Instances are complete as per the defined scope, though annotation limitations apply.

**Are relationships between individual instances made explicit?**
Yes, instances that come from the same participant are linked through a shared ResponseId.

**Are there recommended data splits?**
No.

**Are there any errors, sources of noise, or redundancies?**
Yes, since the dataset is based on voluntarily donated private chats, the format and structure of the messages vary across participants.

**Is the dataset self-contained, or does it link to external resources?**
The dataset is self-contained.

**Does the dataset contain data that might be considered confidential?**
The dataset has been anonymized to remove identifiable information; however, due to the unstructured and varied nature of the donated chats, there remains a small possibility that some private or sensitive details could be unintentionally retained.

**Does the dataset contain data that might be offensive, insulting, or threatening?**
Some chats may include uncivil or offensive language.

### A.3 Collection Process

**How was the data associated with each instance acquired?**
The data was collected through voluntary donations from participants, who shared segments of their private instant messaging conversations.

**What mechanisms were used to collect the data?**
The data was collected via an online survey, where participants were asked to upload excerpts from their private instant messaging conversations and answer questions about their demographics, relationship with each chat partner, and messaging habits.

**If the dataset is a sample, what was the sampling strategy?**
The dataset is not a sample

**Who was involved in the data collection process?**
Researchers at <Anonymous university> in Singapore.

**Over what timeframe was the data collected?**
January - April 2025

**Were any ethical review processes conducted?**
Yes, with anonymization and responsible data handling protocols.

### A.4 Preprocessing/Cleaning/Labeling

**Was any preprocessing/cleaning/labeling of the data done?**
Yes, we applied standard cleaning steps (e.g., removing URLs, timestamps, normalizing punctuation) and labeled each word with its language and script. Post-processing was used to correct tagging errors and ensure alignment between the original and tagged text.

**Was the raw data saved in addition to the preprocessed data?**
Yes.

**Is the software that was used to preprocess the data available?**
The code for annotation and analysis is documented but not publicly shared.

### A.5 Uses

**Has the dataset been used for any tasks already?**
Yes, the dataset has been used to analyze how code-mixing varies across different levels of relationship intimacy in private instant messaging conversations, as well as language tagging performance.

**Is there a repository linking to papers or systems that use the dataset?**
Yes, available at Anonymous.

**What (other) tasks could the dataset be used for?**

- Studying of code-mixing
- Sociolinguistic analysis of digital communication
- Relationship classification based on the message content

**Is there anything about the composition of the dataset that might impact future uses?**
No.

**Are there tasks for which the dataset should not be used?**
It is unsuitable for personal identification or user profiling.

### A.6 Distribution

**How will the dataset be distributed?**
Publicly accessible at Anonymous.

**What license is the dataset distributed under?**
Creative Commons Attribution 4.0 International (CC BY 4.0).

### A.7 Maintenance

**Who will be supporting/maintaining the dataset?**
The authors.

**Will the dataset be updated?**
Yes, we are planning to release yearly updates.

## B Supplementary Data

### B.1 Data preprocessing

We processed our code-mixed chats by first performing standard cleaning steps, including:

- Removing urls, emails, file_extensions
- Removing dates and timestamps
- Expanding common contractions
- Removing extra spaces and punctuations (except !, ?, and >)

In addition, we added spaces between words in different scripts for cleaner language identification. We also did not lowercase text or remove emojis and numbers to facilitate better identification of proper nouns.

### B.1.1 Identifying Different Turns in a Conversation

After cleaning each conversation, we break it down into a list of message turns. Since conversations were sourced from different platforms, there were differences in how messages were demarcated, but most instances were covered by the following two methods:

- Splitting messages after the newline character.
- Splitting messages after the > character.

For the few cases where there were no discernible patterns to demarcate messages, the newline character was taken as a proxy measure.

### B.1.2 Identifying Speaker Names

From these lists of messages, we identified and extracted the speaker names associated with each message in a conversation. This was done by detecting frequently-occurring words and phrases for each conversation, which were used to detect possible usernames occurring at the start of each message. Slight modifications were made to this technique depending on the conversation length. For example, common phrases would need to satisfy stricter criteria if there are limited messages in the conversation. Finally, some conversations had no mention of the usernames, so these methods were not applicable.

### B.1.3 Language Identification

Language identification in code-mixing contexts, where multiple languages occured in a single sentence, is a non-trivial problem.

We began with rule-based methods to classify words as either English or non-English. These included:

- NLTK's WordNet interface, which provides access to a large lexical database of English words and their relationships
- wordfreq Python library, which offers frequency statistics

We also evaluated model-based approaches that allowed us to tag languages beyond just English and non-English, including:

- FastText's pretrained language identification model
- papluca/xlm-roberta-base-language-detection model, a multilingual XLM-RoBERTa-based classifier fune-tuned for language detection (Papluca, 2020).

While rule-based and pretrained model-based methods provided a useful baseline, their lack of flexibility and accuracy in capturing the nuances of informal, code-mixed texts—particularly in Southeast Asian online communication—motivated our next approach. Given these limitations, our next step involved applying token-level language identification of the cleaned data. We employed the Qwen2.5-14B-Instruct large-language model for this task due to its open source nature, large context window, and the ability to work with Singlish languages (Hui et al., 2024).

- The input pipeline was a list of messages across conversations, with excessively long messages chunked into smaller messages.
- The model was prompted to focus on Southeast Asian languages, and account for online texting patterns. The complete prompt has been provided in the Appendix.
- The output was formatted as a list of tuples for each input message, with the tuple being *(word, language)*.

The process was performed for all conversations in the corpus on a NVIDIA H100 80GB HBM3 GPU, with model parameters set as *(max_new_tokens = 2048, temperature = 0.2, batch_size = 16, random_seed = 42)*.

### B.1.4 Cleaning Tags & Script Identification

Certain post-processing steps were applied to reduce any noise in the output generated. Alignment between input words and the words tagged was checked using functions to detect whether:

- certain words were tagged repeatedly
- certain words were not tagged
- certain tagged words were spelled differently
- certain additional words were tagged

These errors were fixed in post-processing for most cases. For a small number of instances where post-processing was not viable (like large number of words untagged), the messages were LLM-tagged again.

The language tags were also analyzed and mapped to a concise list of language tags. For example, this included mapping tags like *(english, eng, en) –> english*. Finally, the script of each word was also detected using regex patterns to detect scripts like Roman, Chinese, Korean, Devanagari, etc. The final output for each input word was a triplet containing *(word, language, script)*.

### B.2 Translation

The entire code-mixed corpus was also translated to English for further analysis. This was also done through the Qwen2.5-14B-Instruct model because of the advantages outlined before. For translation:

- The input pipeline consisted of entire conversations as opposed to messages.
- Initially, translation was attempted at the message level, but many outputs explicitly mentioned lack of sufficient context.
- Finally the corpus was translated at the conversation level to provide extensive context (Larger conversations were chunked into smaller pieces).

The complete prompt can be referred to in the Appendix. The translations were checked and post-processed in case of possible issues. This included correcting cases where certain words were repeated excessively, multiple words were joined together, or certain words were left untranslated.

## C LLM Prompts

The computations were performed on a NVIDIA H100 80GB HBM3 GPU, with model parameters set as *(max_new_tokens = 2048, temperature = 0.05, repetition_penalty=1.2, stop_strings=["\n"], random_seed = 42)*.
The following are the prompts passed to Qwen2.5-14B-Instruct :

**Context passed for Language Tagging**

You are a helpful assistant. You are a language-annotater for text which contains words from English/Hokkien/Mandarin/-Malay/Tamil/ Korean/Japanese/Vietnamese/Hindi. You will serve as an annotator for a dataset. You will be given text and your goal is to tag the specific langauge of each word in it. The text might contain words from English, Hokkien, Mandarin, Malay, Vietnamese, Japanese, Korean, Hindi and Tamil. You will need to tag each word with the language it belongs to. The text might contain lingo or abbreviations from English, Hokkien, Mandarin, Malay, Vietnamese, Japanese, Korean, Hindi and Tamil. You have to tag them in the language they belong to. If you feel the word is an explicit profanity, still tag it in the language it belongs to. Your main goal should be to tag the words in the specified languages, but you can also tag the words in other languages if you feel they are present. This should be done with caution.
The text may contain personal details like names, locations, online usernames, and other proper nouns. Make sure you tag them using the following tuple format: ('word', 'PROPERNOUN'). I do not want any explanations. I want you to simply tag the text without any additional information. In case you encounter any emojis, you can tag them as ('word', 'EMOJI'). In case you encounter any numbers or punctuations, you can tag them as ('word', 'NUMERIC') or ('word', 'PUNCTUATION') respectively. Make sure you tag proper nouns, speaker names, and other 'names' as ('word', 'PROPERNOUN'). Anonymization is key. You should return the output as a list of tuples where each tuple contains the original word and the language it belongs to. The output format should be like this: [ ('word1', 'language1'), ('word2', 'language2'), ... ]. It should be strictly in this format. Be very careful when deciding the tag between Hokkien, Mandarin or Malay. If you are not sure about the language, give your best guess from English, Hokkien, Mandarin, Malay, Tamil, Vietnamese, Japanese, Korean, European, Latin, Bengali, Hindi, Urdu, PROPERNOUN or Other. Strictly tag each and every word in the text. Each word should be outputted in its original form.
Tag the words of the following text and return the list of tuples. Remember the tuples should contain strings only. Do not generate any additional output.

**Context passed for Translation**

You are a helpful assistant. You are a translator from Singlish (a mixture of English/Chinese/Malay/Tamil) to English. You will serve as an annotator for a dataset.
RULES:
You will be given text in Singlish and your goal is to translate it into English. If the entire text is already in English, then keep it in English. If the entire text is partly in English, translate the non-English parts and convert the whole text into English while retaining the meaning. If the entire text is fully in Chinese, Malay, or Tamil, then convert the entire text into English. If some parts of the text are English lingo or abbreviations, convert them into standard English. In case you are unable to translate a word or phrase, return it as it is, but do not leave it blank. The text may contain dates, times, names, locations, online usernames, and other proper nouns. Make sure you anonymize them using the following format: [NAME], [LOCATION], [DATE], [TIME] etc. The text may contain words from other languages like Hindi, Korean, Japanese, etc. Translate them into English also. Make sure to retain the meaning and the nuance of the entire text for all cases. Retain the the tone. Retain incivility. I do not want any explanations. Give a single translation for each word or phrase. I want you to simply translate the text without any additional information. If the entire text is already in English, do not try to make it better. Strictly do not return any notes or comments. Do not leave it blank.
OUTPUT FORMAT:
Translate the following text From Singlish to English and return in string format. Do not return the prompt or any other information, just return strictly translated text. In case of empty string, return empty string.

The final prompts were created in the *chat_template* format by using the following method:

**Prompt Creation**

```
def build_qwen_prompt(
    input_sentence, context,
    tokenizer):
    messages = [
        {"role": "system", "content": context},
        {"role": "user", "content": f"Text: {input_sentence}"}
    ]
    prompt = tokenizer.apply_chat_template(
        messages, tokenize=False,
        add_generation_prompt=True
    )
    return prompt
```

## D Content Analysis

In this section, we examine the role of code-mixing as a conversational resource for managing social interaction. While our previous analyses documented the prevalence of code-mixing in the Singaporean student context (Androutsopoulos, 2006; Zappavigna, 2012), less is known about how it operates across different levels of relational intimacy and conversational goals in everyday messaging. To address this, we present a three-level analysis that moves from statistical patterns to linguistic features and finally to social functions:

- **Level 1: Statistical Patterns.** We begin by quantifying how the frequency of code-mixing varies across different levels of relationship intimacy and message lengths. This provides a descriptive foundation for understanding where code-mixing tends to occur.
- **Level 2: Linguistic Associations.** We then analyze how code-mixing correlates with linguistic features captured by LIWC categories, revealing associations with affect, social processes, and cognitive functions.
- **Level 3: Social Functions.** Finally, we draw on Systemic Functional Linguistics (SFL) to interpret how code-mixing contributes to conversational strategies such as expressing closeness, providing support, or managing affect in context.

### D.1 Level 1 - Statistical Patterns

We begin by characterizing the general structural and stylistic patterns observed in our message-based corpus. Singapore has four official languages: English, Chinese, Malay, and Tamil. 100% of participants used English, either alone or in combination with other languages. While Chinese was the second most commonly used language, Malay and Tamil were used by only 19.88% and 2.4% of participants, respectively. More than half of the participants (51.8%) reported using a combination of just English and Chinese, while Chinese being paired with other languages (such as Malay and others) by nearly 81% of participants. Only 9% of participants communicated exclusively in English, highlighting the widespread use of code-mixing in Singapore. The most frequently used tokens are reported in Figure 3, where words are sized according to their term frequency relative to the others tagged with the same language, and shaded according to their popularity across the chats. The most frequent words in our dataset are dominated by personal pronouns, discourse particles, and colloquial expressions, reflecting the interpersonal and informal nature of everyday messaging. For instance, English tokens such as *ok*, *like*, and *lol* are more prevalent than formal content words typically found in news or academic corpora.

### D.2 Level 2 - Linguistic Associations

Having established where code-mixing is most prevalent across relationship types and message lengths, we next examine how code-mixing relates to the linguistic style and content of conversational messages. Specifically, we investigate whether code-mixed messages tend to co-occur with particular psycholinguistic features that characterize informal, expressive, or socially-oriented discourse.

To do this, we computed Pearson correlations between the percentage of code-mixed tokens and linguistic features derived from the Linguistic Inquiry and Word Count (LIWC) framework. To reduce dimensionality and group related features meaningfully, we performed a principal component analysis (PCA) on the LIWC features and retained five broad categories: *Linguistic Structures*, *Expressive Emotions*, *Social Referencing*, *Cognitive Processes*, and *Contextual Framing*. We report correlation estimates with 95 confidence intervals, assessed using Fisher's r-to-z transformation. Full definitions of the LIWC categories and PCA outcomes are provided in the supplementary materials.

As shown in Figure 5, code-mixing co-occurs with conversational, expressive, and socially-referential language styles, particularly in informal, personal, or emotionally charged contexts. Structural markers (pronouns, function words), expressive devices (punctuation), and social references are reliable linguistic correlates of code-mixed discourse. Key findings include:

- **Linguistic Structure**: Features such as linguistic ($r = 0.26$, $p_{adj} < .001$), function words ($r = 0.28$, $p_{adj} < .001$), and auxiliary verbs ($r = 0.19$, $p_{adj} < .001$) indicate that code-mixed messages tend to use highly interactive and relational language, characterized by pronouns, auxiliary constructions, and discourse particles typical of casual, present-focused conversation.
- **Social Referencing**: The positive associations with pronoun usage (*pronoun*, $r = 0.17$, $p_{adj} = .009$) and female references (*female*, $r = 0.17$, $p_{adj} = .005$) suggest that code-mixing is more frequent in socially engaging discourse, particularly when referencing other people or relational identities.
- **Contextual Framing**: The use of articles (*article*, $r = 0.16$, $p_{adj} < .01$) and determiners (*det*, $r = 0.23$, $p_{adj} < .001$) highlights the role of referential framing, where speakers specify or differentiate entities in context, potentially to manage clarity in multilingual interaction.

Together, these findings suggest that code-mixing is both, a structural feature of multilingual messaging, and a stylistic resource that aligns with conversational goals such as managing social relationships, expressing affect, and negotiating meaning. In the next section, we extend this analysis by applying the SFL framework to interpret how these linguistic patterns contribute to broader conversational strategies and social action.

### D.3 Level 3 - Social Functions

We further apply Halliday's Systemic Functional Linguistics (SFL) framework to interpret how code-mixing functions within the social dynamics of conversation (Halliday and Matthiessen, 2013; Eggins, 2004). SFL describes communication in terms of field, tenor, and mode, which together situate linguistic choices concerning social action. Here, field refers to the type of activity being performed—such as coordinating plans, offering support, or engaging in humor.Tenor captures the relationship between speakers, including their level of intimacy and affective stance. Mode describes the medium and style of interaction, in this case, informal text-based messaging between friends (Matthiessen et al., 2010). Therefore, this framework complements our earlier statistical and linguistic findings by explaining how code-mixing supports specific conversational strategies, accounting for the presence of code-mixing across intimacy levels (Level 1) and its association with linguistic features (Level 2) in how these patterns function socially to manage relationships and actions in everyday conversation (Level 3).

Table 7: Latent Factors, Descriptions, and Dominant LIWC Features

| Label | Definition | Dominant LIWC Features (Loading) |
|---|---|---|
| Linguistic Structures | Features reflecting structural and grammatical complexity of language. | Analytic (−0.83), Linguistic (0.84), function (0.86), pronoun (0.69), ppron (0.60), i (0.60), auxverb (0.67), verb (0.62), focuspresent (0.60), conj (0.35) |
| Expressive Emotions | Features reflecting emotional expression and affective engagement. | emotion (0.85), emo_pos (0.81), filler (0.81), AllPunc (0.68), leisure (0.77), Lifestyle (0.50) |
| Social Referencing | Features reflecting references to social actors and interpersonal relations. | Clout (0.62), pronoun (0.43), ppron (0.55), you (0.46), shehe (0.50), Social (0.72), socrefs (0.81), family (0.43), female (0.53), male (0.37) |
| Cognitive Processes | Features reflecting cognitive mechanisms and references to relational or temporal elements. | Drives (0.65), affiliation (0.60), cogproc (0.53), negate (0.48), cogproc (0.47), differ (0.41) |
| Contextual Framing | Features reflecting contextual elaboration, specificity, and framing of information. | BigWords (0.38), det (−0.52), article (−0.49), Perception (0.55), prep (0.41), motion (0.61), space (0.68), time (0.46) |

While field and mode are useful for capturing the activity type and medium-specific features of digital chat, tenor offers the most analytically rich lens for unpacking the social meaning of code-mixed expression in our data. This is particularly important in computer-mediated communication (CMC), where paralinguistic cues such as intonation, facial expression, and gesture are absent, and relational tone must instead be constructed through text and style.

In our study, we explicitly gathered information about the chat participants, their relationship to each other, and the context of their interaction. Participants were asked how they knew the person they were chatting with, how long they had known them, and to rate their level of intimacy using a set of statements related to trust and emotional disclosure (see Figure 1). This data provides a foundation for analyzing how speakers manage relational intimacy. In this context, tenor is particularly useful. Tenor, as conceptualized by Elena (2016), consists of two interrelated dimensions: **personal tenor** and **functional tenor**. **Personal tenor** refers to the speaker's subjective orientation toward the addressee - whether formal or informal, distant or intimate, polite or direct. It reflects how relationships *feel*. **Functional tenor** concerns the communicative role or the goal of the speaker in a given interaction. It governs how language is used to *do something socially*, such as softening a request, dramatizing a situation, or managing turn-taking.

In what follows, we outline three tenor-based strategies – (a) affective, (b) interpersonal-social, and (c) discourse-stylistic – that speakers deploy to navigate varying levels of closeness, perform emotional or cultural belonging, and construct the tone of interaction in digitally mediated conversation. These strategies are consistent with insights drawn from participants' personal reflections, which reveal how intimacy, social belonging, and cultural background shape their communicative choices (see Table **??**):

- **Affective Tenor:** Affective tenor refers to how speakers express and manage emotions. This strategy was most evident in close, emotionally-invested friendships, where participants were more open in disclosing emo-tions. Within our data, affective tenor served two core functions. The first involves the expression of affection and care, drawing on personal tenor to reflect the speaker's emotional stance toward the addressee and the perceived closeness of the relationship. Code-mixed affection markers such as *sayang, jiayou, thanku* were commonly used to comfort, encourage, or acknowledge shared difficulties. These expressions functioned as clear indicators of relational closeness and were used to signal emotional availability, mutual understanding, and trust. The second function of affective tenor relates more closely to functional tenor, where speakers used code-mixed expressions to modulate emotional intensity and shape the affective tone of the interaction. Expressions such as *wah, jialat, pengsan, or sian* were used to dramatize emotional responses, express fatigue, or inject levity. For example, *"Wah I really cannot than school already"* amplifies stress for effect, while *"relax lah"* serve to downplay frustration. These expressions act as contextualization cues that guide how the utterance is emotionally interpreted by the recipient(Gumperz, 1982). In this way, affective tenor helps regulate emotional meaning in interaction.

Table 8: Descriptions of Dominant LIWC-22 Features Grouped by Latent Category

| Category | Feature | LIWC-22 Definition |
|---|---|---|
| **Linguistic Structure** | Analytic | Degree to which language reflects formal, logical, and hierarchical thinking. |
| | Linguistic | Proportion of dictionary words in the text that match any LIWC dictionary category. |
| | function | Function words such as articles, pronouns, prepositions, auxiliary verbs, etc. |
| | ppron | Personal pronouns (I, you, he, she, we, they). |
| | i | First-person singular pronouns (I, me, my). |
| | auxverb | Auxiliary verbs (am, will, have, been, etc.). |
| | verb | All verb forms excluding auxiliary verbs. |
| | focuspresent | Words referencing the present (is, now, today). |
| | conj | Conjunctions (and, but, whereas). |
| **Expressive Emotion** | emotion | All affective or emotional words. |
| | emo_pos | Positive emotion words (happy, love, excellent). |
| | filler | Filler words or hesitation markers (uh, um). |
| | AllPunc | Total punctuation marks (periods, commas, etc.). |
| | leisure | Leisure-related words (music, movie, party). |
| | Lifestyle | Lifestyle-related words referring to habits or daily activities. |
| **Social Referencing** | Clout | Relative social status, confidence, or leadership conveyed in the text. |
| | pronoun | All pronouns (I, you, he, she, it, we, they). |
| | you | Second-person pronouns (you, your, yours). |
| | shehe | Third-person singular pronouns (he, she, him, her). |
| | Social | Words referring to social processes and relationships (mate, friend, talk). |
| | socrefs | Social references such as friend, buddy, coworker. |
| | family | Family-related words (mother, brother, cousin). |
| | female | Female references (woman, girl, she). |
| | male | Male references (man, boy, he). |
| **Cognitive and Relational Processes** | Drives | Motivational and drive-related words (want, need, desire). |
| | affiliation | Words related to affiliation and belonging (ally, friend). |
| | negate | Negations (no, not, never). |
| | cogproc | Cognitive process words (cause, know, ought). |
| | differ | Words reflecting differentiation or contrast (but, else, although). |
| | discrep | Discrepancy words indicating inconsistency (should, would, could). |
| **Contextual Framing** | BigWords | Words with more than six letters. |
| | det | Determiners (a, an, the). |
| | article | Articles (a, an, the). |
| | Perception | Words related to sensory and perceptual processes (see, hear, feel). |
| | motion | Words describing movement (go, move, run). |
| | space | Spatial references (up, down, above). |
| | time | Temporal references (hour, day, year). |
| | prep | Prepositions (to, with, above). |

- **Interpersonal-Social Tenor:** Interpersonal-social tenor refers to how speakers manage interpersonal relationships and social distance. This was most commonly observed in casual or moderately close relationships, where speakers aimed to maintain rapport without explicitly expressing strong emotion. This strategy draws on both personal and functional tenor: personal, in how speakers position themselves in relation to others; and functional, in how language is used to perform politeness, maintain harmony, or soften interaction. The first core function involves managing closeness, face, and solidarity, often through expressions like *should be ok one, don't worry la, paiseh*. These phrases helped maintain a supportive tone without invoking strong emotion, and reflected culturally familiar politeness strategies aimed at reducing imposition and affirming goodwill (Brown, 1987; Locher and Graham, 2010). This aligns with what prior studies describe as interpersonal and relational dynamics, where relational bonds are managed through small, affiliative moves (Malovana and Yusiuk, 2020). The second function centres on constructing identity and relational softening, achieved through code-mixed discourse particles such as *leh, bah, eh, and liao*. These stylistic markers were used to shape the interactional rhythm, create informality, and soften statements. Utterances like *"U da best leh"* or *"Can liao"* helped speakers express friendliness while keeping the tone light and non-intrusive. Beyond tone management, these expressions also functioned as indexical cues, signalling social identity and peer group affiliation through shared cultural vocabulary. In this way, interpersonal-social tenor enables speakers to fine-tune tone and social alignment, especially in relationships where rapport exists but emotional closeness is low.

- **Discourse-Stylistic Tenor:** Discourse-stylistic tenor refers to how speakers manage the structure and style of conversation. This was observed in both casual and close friendships. This strategy draws primarily on functional tenor, as speakers use linguistic features to coordinate interaction, clarify intent, and maintain conversational flow. It also intersects with personal tenor when style is used to signal peer alignment, cultural familiarity, or in-group belonging. The first core function of discourse-stylistic tenor relates to discourse management. This includes framing, elaboration, justification, and signaling turn-taking or topic shifts — often achieved through discourse particles like *hor, ah, meh, or bah* to guide interpretation or clarify intent. For example, *"Actually ah, since we're talking – you know, anyone looking for job?"* uses *ah* to frame an upcoming request as a casual continuation to the conversation, subtly providing justification for the topic shift. Such markers reflect what (Gumperz, 1982) describes as cognitive and content-related dimensions of discourse, helping speakers build shared understanding, manage relational tone, and justify communicative intent.
The second function — typographic play and stylistic expression, was particularly salient in both casual and close friendships, where speakers used exaggerated phrases like *walao, nani, or siao ah* to dramatise reactions, inject humour, or align affectively with their peers. These expressions reflect what (Androutsopoulos, 2006; Zappavigna, 2012) identify as interactional routines specific to CMC - the patterned, repetitive ways of performing stance and managing interaction. Simultaneously, they function as stylistic identity markers, drawing on pop culture, regional slang, or youth vernacular to stylize voice and signal in-group belonging. Through such choices, speakers construct identity and peer alignment, using stylistic and linguistic features to shape tone and manage the conversation in ways that reflect the social closeness and character of the interaction.